\documentclass[sigconf]{acmart}

\renewcommand\footnotetextcopyrightpermission[1]{} % removes footnote with conference information in first column
\usepackage{booktabs} % For formal tables
\usepackage{amsmath}
\usepackage{blindtext}
\usepackage{graphicx}
\usepackage{subfig}
\usepackage{tabularx}
\usepackage{amssymb}
\usepackage{algorithm}
\usepackage[noend]{algpseudocode}
\usepackage{multirow}

\usepackage{amsmath}

\usepackage{geometry}
\usepackage{algorithm}
\usepackage{algpseudocode}
\usepackage{stfloats}
\usepackage{physics}

\usepackage{amsmath,amsfonts,amssymb}
\usepackage{mathtools}

\usepackage{rotating}

\begin{document}
\title{\vspace{2mm}NESTA: Hamming Weight Compression-Based Neural Proc. Engine\vspace{2mm}}

\author{Ali Mirzaeian, Houman Homayoun, Avesta Sasan \\  George Mason University, Fairfax, VA, USA \\ amirzaei@gmu.edu, hhomayoun@ucdavis.edu, asasan@gmu.edu \vspace{8mm}}

\begin{abstract}
In this paper, we present NESTA, a specialized Neural engine that significantly accelerates the computation of convolution layers in a deep convolutional neural network, while reducing the computational energy. NESTA reformats Convolutions into $3 \times 3$ batches and uses a hierarchy of Hamming Weight Compressors to process each batch. Besides, when processing the convolution across multiple channels, NESTA, rather than computing the precise result of a convolution per channel, quickly computes an approximation of its partial sum, and a residual value such that if added to the approximate partial sum, generates the accurate output. Then, instead of immediately adding the residual, it uses (consumes) the residual when processing the next batch in the hamming weight compressors with available capacity. This mechanism shortens the critical path by avoiding the need to propagate carry signals during each round of computation and speeds up the convolution of each channel. In the last stage of computation, when the partial sum of the last channel is computed, NESTA terminates by adding the residual bits to the approximate output to generate a correct result.
\end{abstract}


\maketitle

% \begin{IEEEkeywords}
% component, formatting, style, styling, insert
% \end{IEEEkeywords}
\section{Introduction and Background} \label{intro}
Deep learning models that deploy Convolutional Neural Networks (CNN) for feature extraction have become increasingly popular in recent years \cite{deep_learning}. The popularity of these learning solutions stems from their ability to achieve unprecedented accuracy, surpassing that of human's ability, for various tasks such as object and scene recognition \cite{alexnet,vgg, googlenet,resnet,icnn, 8697497}, object detection, and object localization\cite{girshick2014rich, sermanet2013overfeat}. This, as illustrated in Table \ref{related_table}, is made possible by using deep and complex neural networks expressed using specialized frameworks such as Caffe \cite{caffe}, PyTorch \cite{ketkar2017introduction} and Tensorflow \cite{abadi2016tensorflow}, and trained and executed in acceptable time by Graphical Processing Units (GPU).  

Although innovation in parallel computing has enabled us to train and execute such complex models, the applicability of such models remains limited due to their computational and storage requirements. These state of the art CNNs require up to hundreds of megabytes for a model and partial result storage and 30k-600k operations per input pixel \cite{eyeris}. The high computational complexity of these models, in turn, poses energy (power) and throughput (delay) challenges to the underlying hardware. Typically, in such learning solutions the majority (over 90\%) of computational complexity is for processing the convolution (CONV) layers \cite{cong2014minimizing}.   

The generality of a processing engine significantly affects the throughput and energy efficiency of neural processing hardware\cite{sankaradas2009massively}\cite{du2015shidiannao}. The more general the hardware, the less efficient (in terms of delay and power) the computation becomes. The least attractive solutions are generated by running CNNs on general-purpose CPUs. Utilizing more specialized hardware such as GPUs and FPGAs provide a significant improvement in the efficiency of computation, while most efficient computing, with an order(s) of magnitude improvement in performance and power consumption, is reported when specialized ASIC accelerators such as Eyeriss\cite{eyeris}, Diannao\cite{diannao}, Dadiannao\cite{dadiannao}, or Shidiannao\cite{peemen2013memory} are deployed. The major difference in the performance of ASIC accelerator solutions, previously proposed in \cite{eyeris, diannao, dadiannao, peemen2013memory, sankaradas2009massively, du2015shidiannao, 8392465, Jafariglsvlsi},
is on the type of data flow implemented for maximizing data reuse (weight, partial sum, and activation value) and minimizing memory access. This is when the neural Processing Elements (PE), that implement the multiply-accumulate (MAC) function, remain non-optimized in these accelerator solutions.

\begin{table}[t]
\centering
\caption{Depth and complexity of some of the existing and modern CNN solutions for object detection.}
\vspace{-4mm}
\label{related_table}
\scalebox{0.85}{
    \begin{tabular}{| c | c c c c | } \hline
        &  AlexNet\cite{alexnet} & VGG\cite{vgg} &   GoogLeNet\cite{googlenet} & Resnet\cite{resnet}\\ \hline
        Top5 Accuracy & 80.2\% &    89.6\% & 89.9\% & 96.3\% \\
        layers & 8 &19 & 22 & 152\\
        %$3 \times 3$ Conv Layers & 3 & {\color{red}14} & {\color{red} 10} & {\color{red}50} \\
        FLOPS & 729M &  19.6G  & 1.5G & 11.3G\\
        FLOPS in $3\times3$ CONV & 118M & 19.5G  & 1.18G & 6.7G \\     
        \hline     
    \end{tabular}
}
\normalsize
\vspace{-4mm}
\end{table}

In this paper, we claim that the architecture of PEs in an ASIC DNN accelerator could significantly improve when the computational model, data locality, and data reuse concepts are used to architect a DNN/CNN specific PE. We propose NESTA as a PE that is designed based on these principles. To reduce data movement, and minimize the generation of partial sums, NESTA consumes 9 values of the convolution at a time (equal to the size of a 3 $\times$ 3 convolution) until all filter-image pairs of a convolution across all channels are consumed. To significantly speed up the computation and reduce energy consumption, NESTA does not use adders or multipliers. Instead, it converts the convolution into a sequence of N compression and one final addition. The add operation transforms the compressed and accumulated result into a correct partial sum. 

 \begin{figure}[htb!]
  \centering
  \includegraphics[width=8cm]{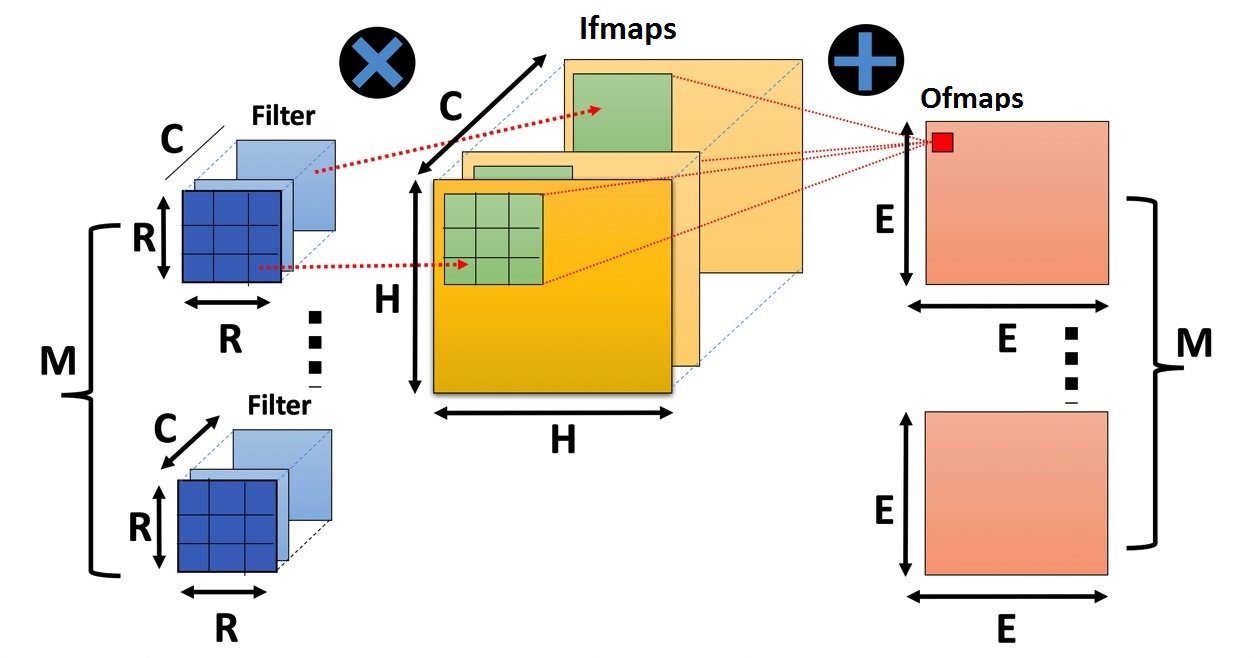}
  \vspace{-4mm}
  \caption [ Computing one CONV layer using input \textit{Ifmap}/image and filters to produce the output \textit{Ofmaps} ] {\textit{\textbf{Computing one CONV layer using input Ifmap/image and filters to produce the output (\textit{Ofmaps)}}} \vspace{-4mm}}
 \label{conv_mm}
 \end{figure}

\begin{figure*}[hbt!]
    \centering
    \includegraphics[width=1.65\columnwidth]{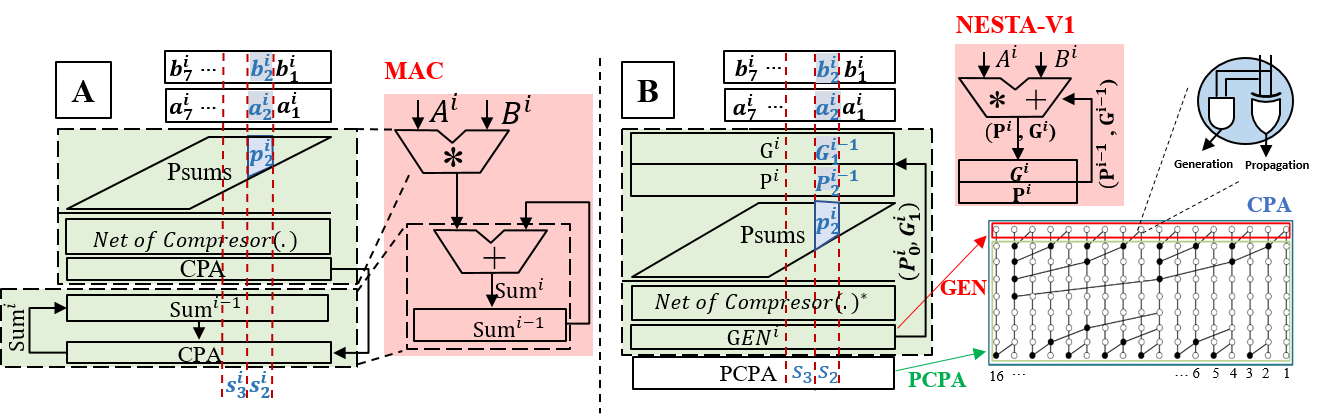}
    \vspace{-5mm}
    \caption{comparing the architecture of A) a typical MAC, versus B) a simplified 2-input version of NESTA. In all variables in the form of $D^i_m$, the subscript ($m$) captures the bit position values, and postscript ($i$) capture the cycle (iteration). For example, $A^i, B^i$ are the input data 
    in the $i^{th}$ iteration (corresponding to the $i^{th}$ cycle) of the multiply accumulate operation. The $b_{m}^{i}, a_{m}^{i}$, and $p_{m}^{i}$ are accordingly the $m^{th}$ significant bits of inputs $A$, $B$, and partial sum at the $i^{th}$ cycle (iteration). The division of CPA into GEN and PCPA is also shown in this figure. Note that the $PCPA$ is only executed at the last cycle.
    \label{GV_NESTA}}
    \vspace{-2mm}
\end{figure*}

\section{NESTA: Proposed Processing Engine}

Before describing our proposed solution, we first explain the concept of \emph{temporal carry} in a miniaturized solution in section \ref{motivation}, then we explain the concept of \emph{compression and expansion} in section \ref{multi_input_HW_Adder}. Finally, in section \ref{nesta_architecture}, we use these concepts to construct and describe our proposed solution.  

\subsection{Motivation 1: Temporal Carry}\label{motivation}

Suppose two vectors $A$ and $B$ each have $N$ 8-bit values, and the goal is to compute their dot product, $\sum_{i=0}^{N-1}(A_i*B_i)$ (similar to what is done during the activation process of each neuron in a NN). This could be accomplished using a single Multiply-Accumulate (MAC) unit and working on 2 inputs at a time for N rounds. Fig. \ref{GV_NESTA}(A-right) shows the General view of a typical MAC architecture that comprised of two parts multiplication and addition. We have assumed that a Carry Propagation Adder (CPA) is used as adder unit for reducing the MAC delay. More detailed view of this architecture, \ref{GV_NESTA}(A-left), reveals that for generating the final result, the CPA will be executed 2N times, N times for producing the results of N multiplications and N times for accumulating the result of multiplications. These CPAs are located at the critical path of this architecture so eliminating them lead to a performance gain. Fig. \ref{GV_NESTA}(A-right) captures how CPA has been broken into GEN (which is highlighted in red), and PCPA (Partial CPA).

Fig.\ref{GV_NESTA}(B-right) shows a simplified version of our proposed solution, NESTA-V1. As illustrated NESTA-V1, 1) intertwines the multiplication and addition, and 2) reduces the delay of CPA by only using the GEN section of the CPA. The GEN section only produces the first level generate $G^i$, and propagate $P^i$ signals, after which NESTA-V1 feedback each $P^i$ and $G^i$ to its compressor network for inclusion in the cycle computation. We can consider this as the process of \emph{generating a temporal carry signal, as opposed to a spatial carry signal} which is used in typical MACs. This is made possible, considering that we do not need the output of individual multiplications, and our target is to compute the correct $\sum_{i=0}^{N-1}(A_i*B_i)$. Hence, in NESTA-V1 for N-1 times, only the GEN section of CPA is executed, while for the last iteration the complete CPA is executed (including PCPA) to avoid generating further temporal carry bits.

\subsection{Motivation 2: Compression and Expansion}\label{multi_input_HW_Adder}
\begin{figure*}
    \centering
    \begin{tabular}{cc}
        \includegraphics[width=0.4\columnwidth]{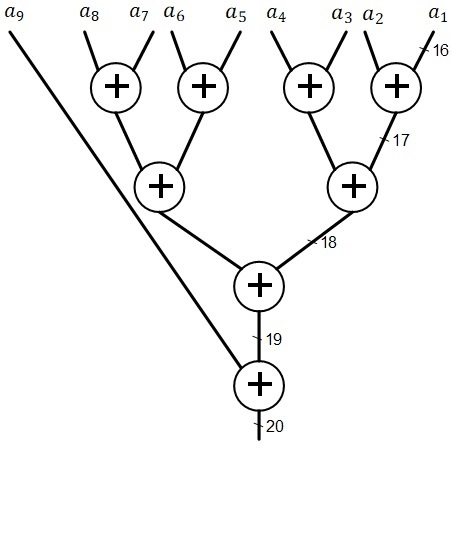} & \includegraphics[width=1.3\columnwidth]{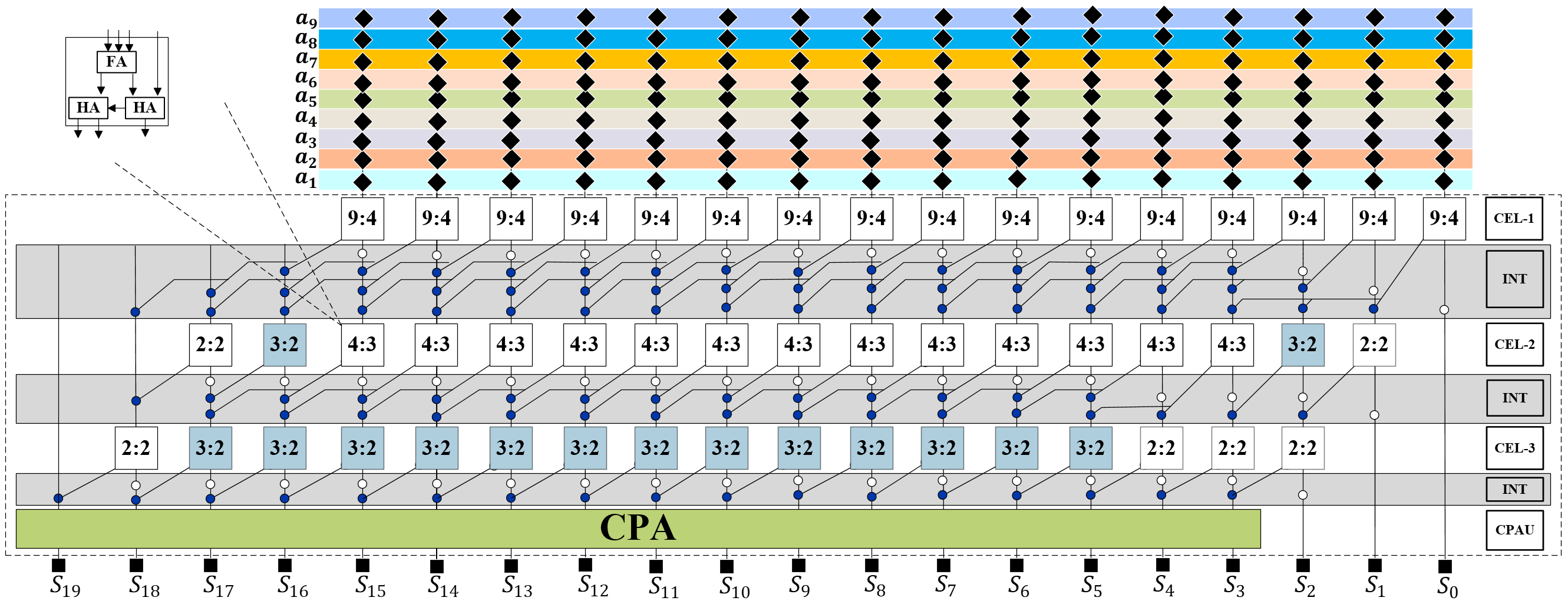}  
    \end{tabular}
    \vspace{-4mm}
    \caption{An Adder tree for 9 16-bit-width entries (left), Hamming Weight Adder (HW-Adder) of 9 16-bit-wide entries (right). In the HWC-Adder compressor hierarchy (CEL) the complete compressors are colored blue, while compressors with available capacity are white. For building the improved version of HWC-Adder (HWC-Adder*), 2 bits from each compressor in CEL-1 are differed to a compressor in the same bit position in CEL-2, increasing the number of complete compressors and reducing the critical path delay in CEL-1 as reported in table \ref{tree-hwa}}
    \label{hamming_weight_adder}
\end{figure*}

Lets consider an application that requires hardware acceleration for computing the following expression: $p = \sum_{i=1}^{9}a_{i}$, in which $a_{i}$(s) are 16-bit unsigned numbers. One natural solution, as illustrated in Fig. \ref{hamming_weight_adder}.(left), is using an adder-tree, while each add operator could be implemented using a fast adder such as carry-look-ahead \cite{kwon2001fast} (CLA),  Brent-Kung \cite{brent1982regular} (BK) or Kogge-Stone \cite{kogge1973parallel} (KS) adder.  Regardless of the choice of the adder, the resulting adder tree is not the most efficient. The adder power delay product (PDP) could significantly improve if a multi-input adder is reconstructed using Hamming Weight (HW) compressors. For this purpose, we reformulate the computation of $p$ as shown in Equation \ref{reformulate_sum}, by rearranging the values into 16 arrays, where each array is composed of 9 bits with equal significance value. With this formulation, we can use a hierarchy of Hamming Weight compressor to perform the addition. 

\vspace{-4 mm}
\begin{equation}\label{reformulate_sum}
    p = \sum_{i=0}^{15}\sum_{j=1}^{9}(2^{i} \And a_{j})
\end{equation}

Fig. \ref{hamming_weight_adder}-(right) captures the structure of the proposed HW compression Adder (HWC-Adder), which is composed of 4 stages. In each of the first 3 stages, the HW compressors C(m:n) take a stack of $m$ bit values of the same significance (shown vertically) and computes its HW value (of size $n$) which is expanded vertically. Aligning the bit values of the same significance generates a smaller stack of bit values at each bit position as input to the next level of compressors. We refer to each of these stages (stages 1 to 3) as Compression and Expansion Layer (CEL). In the last stage, every bit-column contains no more than 2 bits. In this stage, a 2-input addition generates the final results.   

Table \ref{tree-hwa} compares the PPA and PDP of an adder tree constructed using Brent-Kung and Kogge-Stone adders, and that of HWC-Adder. As illustrated the energy consumption of the HWC-Adder is 50.2\% and 39.8\% lower than that of the BK and KS adder-trees respectively. At the same time, the delay of HWC-Adder is 8.3\% and 9.8\% lower than that of the KS and BK adder-trees respectively. The delay of HWC-Adder architecture could be further improved, if instead of incomplete C(9:4) HW compressors in the first CEL, we use complete CC(7:3) compressors, passing the unconsumed bits (2 bits) to the next hierarchy layer, in which the C(4:3) incomplete compressors are converted to C(6:3). This transformation shortens the critical path (reduces the number of logic levels) in stage CEL-1 and reduces the area, without increasing the number of logic levels in CEL-2, hence, producing a faster implementation. The PDP of the resulting HWC-Adder$^*$ is captured in the table \ref{tree-hwa}. The resulting improvements in the HWC-Adder(s) are the result of 1) using larger HW compressors (as opposed to C(2:2) and C(3:2) in Brent-Kung), and 2) maximizing the number of complete compressors, thus reducing the hardware deficiency.  

\begin{table}
\caption{Comparing the efficiency of HWC-Adder(s) vs Adder tree constructed using Brent-Kung (BK) and Kogge-Stone (KS).}
\vspace{-4mm}
\centering
\scalebox{0.9}{
\begin{tabular}{|l|c|c|c|c|}
%\toprule
\hline
\textbf{Adder Type} & \textbf{Area($um^2$)} & \textbf{Delay($ns$)} & \textbf{Power($uW$)} & \textbf{PDP($fj$)}\tabularnewline \hline 
Adder tree (BK)      &   4723      &   2.66 &   0.555  &  1.48  \tabularnewline 
Adder tree (KS)      &   5135      &   2.60 &   0.686  &  1.78  \tabularnewline
HWC-Adder            &   4738      &   2.40 &   0.369  &  0.88  \tabularnewline 
HWC-Adder$^*$        &   4428      &   2.35 &   0.368  &  0.86  \tabularnewline 
\hline
\end{tabular}
}

\label{tree-hwa}
\vspace{-5mm}
\end{table}

\subsection{NESTA: Our Proposed Solution}\label{nesta_architecture}

Our proposed solution, NESTA, is a specialized neural processing engine designed for executing learning models in which filter-weights, input-data, and applied biases are expressed in fixed-point format. NESTA combines 9 multiplications and 9 additions into one batch-operation for gaining energy and performance benefits. Let's assume $NESTA_{ACC}$ is the current accumulated value, while $I$ and $W$ represent the input values and filter weights respectively. In its n$^{th}$ round of execution, NESTA performs the following operation:
\vspace{-2mm}
\begin{equation}
\vspace{-2mm}
   NESTA_{ACC}(n) = NESTA_{ACC}(n-1) + \sum_{i=9n}^{9n+9} I_{i} \times W_{i}
\end{equation}

To improve efficiency, NESTA does not use adders and multipliers. Instead, it uses a sequence of hamming weight compressions followed by a single add operation. Furthermore, in each cycle $c$, after consuming 9 input-pairs (weight and input), instead of computing the correct accumulated sum, NESTA quickly computes an approximate partial sum $S^{'}[c]$ and a carry $C[c]$ such that $S[c] = S^{'}[c] + C[c]$. The $S^{'}[c]$ is the collection of generated bits ($Gi$) and $C[c]$ is the collection of propagated ($Pi$) bits produced by GEN unit of CPA. Note that the division of CPA into GEN and PCPA was described in section \ref{motivation}. The $S^{'}[c]$ is saved in the output registers, while the $C[c]$ are stored in Carry Buffer Unit (CBU) registers. In the next cycle, both $S^{'}[c]$ and $C[c]$ are used as additional inputs (along with 9 new inputs and weights) to the CEL unit. Saving the carry (propagate) values ($P$s) in CBU and using them in the next iteration reflects the temporal carry concept that was described in section \ref{motivation}, while the reuse of $S^{'}$ in the next round implements the accumulation function of NESTA.  

In the last cycle, when working on the last batch of inputs, NESTA computes the correct $S[c]$ by using the PCPA to consume the remaining carry bits and by performing the complete addition $S[c] = S^{'}[c] + C[c]$. Note that the add operation generates a correct partial sum whenever executed. But, to avoid the delay of the add operation, NESTA postpones it until the last cycle. For example, when processing a $11 \times 11$ convolution across 10 channels, to compute each value in Ofmap, 1210 ($11\times11\times10$) MAC operations are needed. To compute this convolution, NESTA is used 135 times $\lceil 1210/9 \rceil$, followed by one single add operation at the end to generate the correct output. 

\begin{figure*}[hbt!]
    \centering
    \includegraphics[width=2\columnwidth]{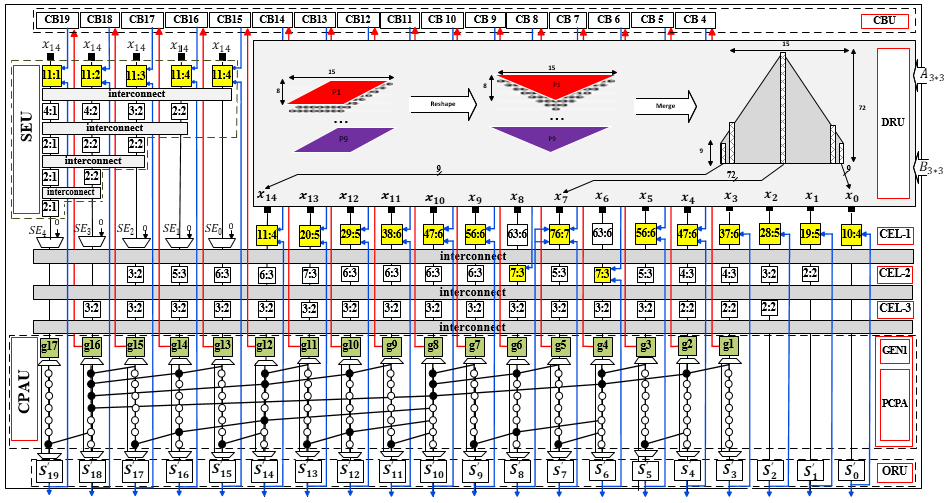}
    \vspace{-4mm}
    \caption{In NESTA carry bits that are generated in GEN section of the CPA do not propagate into the carry chain. Instead, they are captured by CB registers. In the next cycle, the carry bits (of the previous cycle that are stored in CB registers) are fed to the hamming weight compressors at that bit position, temporally deferring their impact to the next cycle. The compression unit, in each cycle consumes the bit values from 9 new input (W, I) pairs, the Carry bits of the previous cycle (stored in CB registers) and the partial sum stored in S registers. The consumption of bit values in S registers implement the accumulation function.  In the last round of computation, instead of capturing the carry bits in CB registers, they are fed to the PCPA  (Partial CPA) to propagate through the carry chain and generate the correct convolution results. }
    \label{hwc_with_grouping}
    \vspace{-4mm}
\end{figure*}

Fig. \ref{hwc_with_grouping} captures the NESTA architecture. It is comprised of 6 units: 1) Data Reshaping Unit (DRU), 2) Sign Expansion Unit (SEU), 3) Compression and Expansion Layers (CEL), 4) Adder Unit (AU), 5) Carry Buffer Unit(CBU), and 6) Output Register Unit(ORU). Each of these units is described next:

\subsubsection{Data Reshape Unit (DRU)} 
The DRU, as illustrated in Fig. \ref{hwc_with_grouping}-(DRU), receives 9 pair of multiplicands and multipliers (W and I), converts each multiplication to a sequence of additions by ANDing each bit value of multiplier with the multiplicand and shifting the resulting binary by the appropriate amount, and returns bit-aligned version of the resulted partial products.

\subsubsection{Sign Extension Unit:(SEU)} The SEU is responsible for producing the sign bits $SE_0$ to $SE_4$. The inputs to SEU is sign bit ($X_{14}$). The result of a multiplying and adding 9, 8-bit values is at most 20-bits. Hence, we need to sign-extend each one of the 15-bit partial sums (for supporting larger the architecture is accordingly modified). To support singed inputs, we also need to slightly change the input data representation. For a partial product $p=a\times b$, if one values $a$ or $b$ is negative, we need to make sure that the negative number is used as the multiplier and the positive one as the multiplicand. With this arrangement, we treat the generated partial sums as positive values and make a correction for this assumption by adding the two's complement of the multiplicand during the last step of generating the partial sum. This feature is built into the architecture using a simple 1-bit sign detection unit, and by adding multiplexers to the output of input registers to capture the sign bits. Note that multiplexers are only needed for the last 5-bits as shown in figure \ref{hwc_with_grouping}-(SEU). Following example clarify this concept: let's suppose that $a$ is a positive and $b$ is a negative b-bit binary. The multiplication $b\times a$ can be reformulated as:

\vspace{-2 mm}
\begin{equation}
   b \times a = (-2^{7}+\sum_{i=0}^{6}x_{i}2^{i}) \times a= -2^{7}a+(\sum_{i=0}^{6}x_{i}2^{i}) \times a
\end{equation}

The term $-2^{7}a$ is the two's complement of multiplicand which is shifted to the left by 7 bits, and the term ($\sum_{i=0}^{6}x_{i}2^{i}) \times a$ is only accumulating shifted version of the multiplicand. Note that some of the output bits generated by SEU compressor extend beyond 20 required bits. These sign bits are safely ignored. Finally, the multiplexers switch at the output of SEU is used to allow NESTA to switch between signed and unsigned modes of operation.

\subsubsection{Compression and Expansion layers (CEL)} 
The input to i$th$ bit of CEL unit in cycle $n$ is the 1) bit-aligned partial sums (at the output of DRU) in position i 2) the temporary sum generated by GEN unit of NESTA at time $c-1$ at bit position $i$, and 3) the Propagate (carry) value generated by GEN unit of NESTA at time $c-1$ at bit position $i-1$. Following the concept of HWC-Adder, described in section \ref{multi_input_HW_Adder}, the CEL is constructed using a network of Hamming Weight Compressors (HWC). A HWC function C$_{HW}$(m:n) is defined as the Hamming Weight (HW) of $m$ input-bits (of the same bit-significance value) which is represented by an $n$-bit binary number, where $n$ is related to $m$ by: $n = \lfloor log_{2}^{m}\rfloor+1$. For example "011010", "111000", and "000111" could be the input to a C$_{HW}$(6:3), and all three inputs generate the same Hamming weight value represented by "011". A Completed HWC function CC$_{HW}$(m:n) is defined as a C$_{HW}$ function, in which  $m$ is $2^{n}-1$ (e.g.,  CC(3:2) or CC(15:4)). As illustrated in Fig.\ref{hwc_with_grouping}, each HWC takes a column of m input bits (of the same significance value) and generate its n-bit hamming weight. The resulting n bits is then horizontally distributed as input to $C_{HW}$(s) in the next-layer CEL. This process is repeated until each column contains no more than 2-bits.

\subsubsection{Carry Propagation Adder Unit(CPAU)} Similar to HWC-Adder, described in section \ref{multi_input_HW_Adder}, the CPA is divided into GEN and PCPA. If NESTA is executed n times, the PCPA is skipped  $n-1$ times and is only executed in the last iteration. GEN is the first logic level of CPA executing the generate and propagate functions to produce temporary sum/generate $G$ and carry/propagate $P$ which are used as input in the next cycle.

\subsubsection{Carry Buffer Unit (CBU)} The CBU is a set of registers that store the propagate/carry bits generated by GEN at each cycle, and provide this value to CEL unit in the next cycle. Note that CB bits can be injected to any of the $C_{HW}(m:n)$ in any of the CEL layers in that bit position. Hence, it is desired to inject the CB bits to an incomplete $C_{HW}(m:n)$ to avoid an increase in the critical path delay of CEL.

\subsubsection{Output Register Unit (ORU)} The ORU capture the output of  GEN in the first n-1 cycles or PCPA in the last cycle of operation. Hence, in the first $n-1$ cycle,  NESTA stores the Generate ($G$) output of GEN unit and feeds this value back to the CEL unit in the next cycle. In the last cycle, it stores the sum generated by PCPA.

\subsection{NESTA: Putting it all together}\label{multicycle}
\begin{figure}[h]
\vspace{-2mm}
    \centering
    \includegraphics[width=0.9\columnwidth]{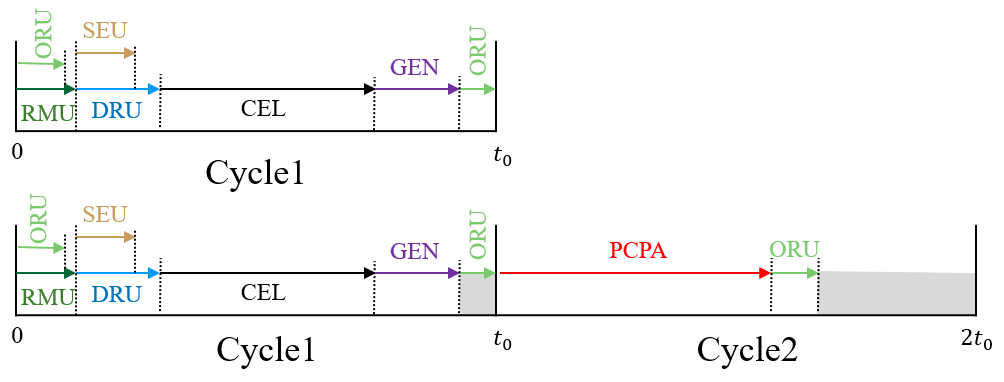}
    \caption{NESTA cycle time is computed by excluding the execution time of PCPA. In the last cycle of computation of convolution, the NESTA activates the PCPA and captured the correct sum after 2 cycles of execution.}
    \label{sequence}
    \vspace{-4 mm}
\end{figure}

NESTA receives 9 pair of Ws and Is. The DRU generate the partial products and bit-align them as input to the CEL unit. The CEL unit at each round of computation consumes 1) bit values generated by DRU, 2) generate (temporary sum) values stored at S registers, and 3) propagate (carry) bits in CB registers. This is when the SEU assures that the sign bits are properly generated. For the first n cycles, only the GEN unit of CPA is executed. This allows NESTA to skip the delay of the carry chain of the PCPA.  To be efficient, the clock period of NESTA is reduced to exclude the time needed for the execution of PCPA. The timing paths in PCPA are defined as multi-cycle paths (2 cycle paths). Hence, the execution of the last cycle of NESTA takes 2 cycles, see Fig. \ref{sequence}. In the last round of execution, the PCPA unit is activated, allowing the addition of stored values in S registers and CB registers to take place for producing the correct and final SUM.  Considering that the number of channels in each layer of modern CNNs is fairly large (128 to 512) the savings in the result of shortening NESTA cycle time (by excluding PCPA) accumulated over large number of cycles (of NESTA execution) is far larger than one additional cycle needed at the end to execute the PCPA for producing the correct final sum.

\begin{figure*}[tb]
    \centering
    \includegraphics[width=1.8\columnwidth]{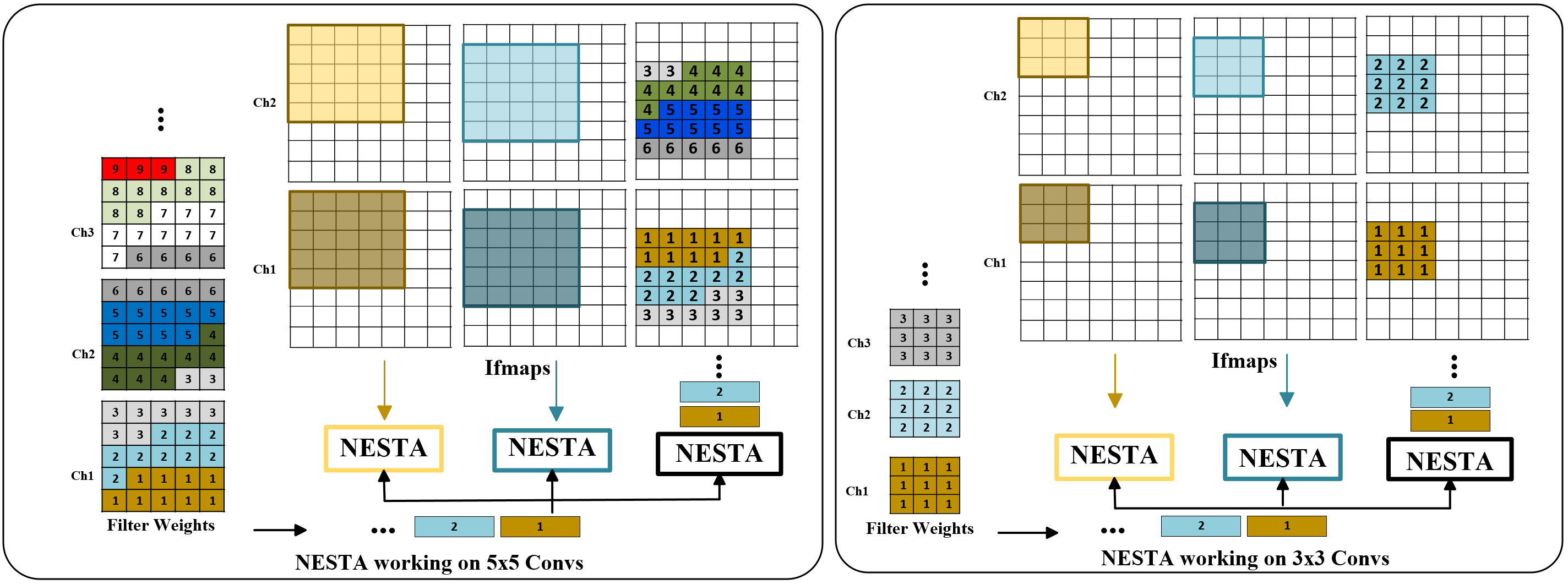}
    \vspace{-2mm}
    \caption{NESTA Row Stationary (RS) data flow for executing $3 \times 3$ convolution across multiple channels (right) and $5 \times 5$ convolution across multiple channels (left). A similar concept can be used to support all other convolutions sizes.}
    \label{RS}
   \vspace{-2 mm}
\end{figure*}

\subsection{Supported Data flows}\label{dataflow_secion}

A considerable portion of the power consumed in a neural processing engine is related to storage, read and write from its memory subsystem. The extent of power consumed in the memory subsystem is a function of 1) the read/write/retention power of each memory element, and 2) the frequency of access to each memory.  In several prior work \cite{sasan1,sasan2,sasan3,sasan4,sasan5,sasan6}, it was shown that it is possible to significantly reduce the  read/write/retention power consumption of a memory unit by aggressively scaling it supplied voltage while deploying architectural fault tolerance techniques and solutions to mitigate the increase in the memory write/read/retention failure rate.  The frequency of access to the memories, on the other hand, can not be controlled from an architectural perspective as it is a dataflow optimization problem.

Memory access pattern of a model which is being executed on a neural engine significantly impacts its energy efficiency and performance. Accessing data in off-chip DRAM consumes around 200X more energy and takes around 20X longer compared to accessing data in on-chip SRAM memories \cite{eyeris}. Hence, for a modern Deep Neural Network with a large number of operations and parameters, designing a dataflow that minimizes the access to off-chip DRAM and maximizes the data reuse (while data is on-chip) can go a long way in improving its energy efficiency and performance. Related to neural processing engines, several dataflows has been studied in the literature. The work in \cite{eyeris} divides the DNN dataflows into 5 major categories: 1) No Local Reuse(NLR), 2) Weight Stationary(WS), 3) Input Stationary(IS), 4) Output Stationary(OS), and 5) Row Stationary(RS, RS+). These data flows differ on the way they reuse input frame maps (Ifmaps), partial sums (Psums), and filter weights. The NLR does not have any reuse at the PE level and requires the largest number of transaction with a global buffer.  Diannao is an example of NLR based accelerator described in \cite{diannao}. The WS dataflow stores the filter weights within the PEs. The goal is to minimize the re-fetching of filter weights by limit their movement. Examples of WS implementation include \cite{Chakradhar:2010:DCC:1816038.1815993}\cite{Gokhale:2014:GMC:2679599.2679649}. IS and OS dataflows try to minimize the movement of Ifmaps and Psums respectively, examples of which include \cite{du2015shidiannao}\cite{gupta2015deep}\cite{peemen2013memory}. The RS dataflow combines the WS and the OS dataflows to achieve greater efficiency. Eyeriss is an example of RS implementation described in \cite{eyeris}\cite{DBLP:journals/corr/abs-1807-07928}. 

Another way to understand the differences between these dataflows is through the study of the algorithm governing the computation of the convolution in each data flow. Let us consider the convolution in Fig. \ref{conv_mm} with M filters (each with size $C \times R \times R$), repeated in a batch of B images with each image being of size $C \times H \times H$. To process this CONV, as shown in Alg. \ref{nested_loop}, seven nested loops are required. Because each one of the loops is independent of the others, changing the order of each these loops can produce a new dataflow. Each dataflow promotes a different form of data reuse. It should be noted that it is possible that one permutation of these nested loops to be applicable to more that one dataflow. For example in the Alg. \ref{nested_loop}, execution line order 1-2-3-4-5-6-7-8, NLR, WS, and RS have the same representation, however, depending on the underlying NOC different data access patterns can be designed. Os and IS dataflows also can be obtained if the execution's line of the seven loops changes to 1-2-4-5-3-6-7. 

\begin{algorithm}
\caption{seven nested loops for calculating an Ofmap. B, M, C, H, R are Batch-size, Number of Filters, Channel size, Height/weight of an ifmap, and filter size respectively.}\label{nested_loop}
\begin{algorithmic}[1]
\footnotesize
\For{$(b=0; b<B; b++)$}
    \For{$(u=0; u<M; u++)$}
        \For{$(c=0; c<C; c++)$}
            \For{$(h=0; h<H; h+=S)$}
                \For{$(w=0; w<H; w+=S)$}  
                    \For{$(i=0; i<R; i++)$}
                        \For{$(j=0; j<R; j++)$}
                            \State ofmap[b][u][h][w] += 
                            \state ifmap[b][c][h+i][w+j]*
                            \State filter[u][c][i][j]
                        \EndFor
                    \EndFor
                \EndFor    
            \EndFor
        \EndFor
    \EndFor
\EndFor

\normalsize
\end{algorithmic}
\end{algorithm}

NESTA could be used to implement any of these dataflows. However, in this work (for lack of space), we only describe how NESTA dataflow could be designed to model the RS dataflow and will address the implementation of other dataflows for our future work. Fig. \ref{RS} capture the RS dataflow used to compute $3 \times 3$ (right) and $5 \times 5$ (left) convolution across many channels. To capture the data reuse and communications between NESTA cores (assuming that many NESTA cores are packed into a SOC to build a many-core accelerator), we have used three NESTAs to construct each of scenarios illustrated in Fig. \ref{RS}. Since NESTA accept 9 inputs at a time, it can perform a $3 \times 3$ convolution in one cycle, or a $5 \times 5$ convolution in 3 cycles. The data is reshaped in the accelerator's global buffer and is streamed to the NESTA cores. Depending on the number of available NESTA cores we can compute the value of different neurons in parallel to promote higher data reuse. In this case we can either 1) compute the neurons in different OFmaps by loading different weights to each group of NESTA and share the ifmap weights (not shown in this figure), or 2) compute the neuron values in the same Ofmap by sharing the weights across different NESTA cores and stream different (partially overlapping) ifmap values to each group as shown in Fig. \ref{RS}.

As described in section \ref{multicycle}, in its last round computations( when working on convolution across multiple channels), NESTA switches to its two-cycle operation mode and activates the PCPA that would take 2 cycles to generate the correct final sum. Note that in deep channels, or for large convolutions, the cost of one extra cycle is negligible compared to the gain of removing the PCPA from the critical path in all computational cycles.

\section{Results}
In this section, we evaluate the NESTA in terms of Power, Performance, and Area (PPA).  NESTA and all MACs cited in this section support 16-bit signed fixed-point inputs.

\subsection{Evaluation and Comparison Framework}\label{framwwork}
\begin{figure*}[!htb]
    \centering
    \includegraphics[width=1.8\columnwidth]{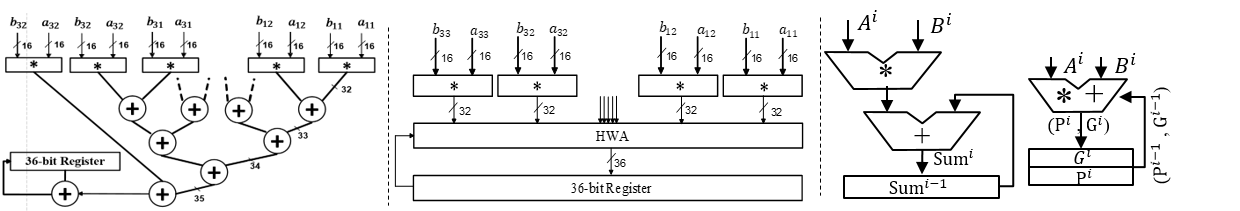}
    \caption{A 9-input MAC, which is identified as (Multiplier Choice, Adder Choice). MACs constructed with similar structure are used for PPA and PDP comparison with our proposed NESTA PE solution. \vspace{-2mm}}
    \label{mac_flavors}
\end{figure*}

The PPA metrics are extracted from the post-layout simulation of each design. Each MAC or MAC9 is designed in VHDL, synthesized using Synopsis Design Compiler \cite{dc} using 32nm standard cell libraries, and is subjected to physical design (targeting max frequency) by using the reference flow provided by Synopsys and by using IC Compiler \cite{icc}. The area and delay metrics are reported using Synopsys Primetime \cite{pt}. The reported power is then averaged across 20K cycles of simulation with random input data fed to PrimetimePX \cite{pt} in FSDB format. To build a fair comparison, in addition to simple 2-input MACs, we constructed multiple flavors of 9-input MACs (MAC9s) using various high-speed adders and multipliers to compute the convolution in one shot. The general structure of MACs and MAC9s used for comparison is captured in Fig. \ref{mac_flavors}. Each MAC9 is constructed using 9 multipliers, the output of which is fed to a 10-input adder tree (9 inputs from multiplier and 1 from output register) to compute a $3 \times 3$ convolution in one shot. For multiplication, we used Booth-Radix-N (BRx2, BRx4, BRx8), and Wallace multipliers. For addition, we used Brent-Kung (BK) and Kogge-Stone (KS) adders. In addition, we considered a hybrid approach, where the multipliers are fed to an HWC-Adder which was discussed in section \ref{multi_input_HW_Adder}.  Each 2-input MAC is identified by (Multiplier choice, Adder choice) and each 9-input MAC9 is identified by (Multiplier Choice, ( Adder Arrangement, Adder Choice)). For example ( BRx2, (tree, Brent-Kung)) is a MAC9 constructed by using 9 BRx2 multipliers followed by an adder tree constructed from Brent-Kung Adders. Similarly, a (BRx2, (HWC-Adder, Brent-Kung)) uses the same multiplier, but replace the adder tree with an HWC-Adder that uses a single Brent-Kung adder.  

\subsection{PPA efficiency: NESTA v.s. MAC9s} \label{CAC_PPA}

\textbf{Power:} The power consumption of NESTA is considerably less than other MAC9 flavors. When comparing NESTA with various flavors of MAC9, the power consumption is reduced by 17.4\% to 58.9\% when compared to (BRX4, (HWA, BK)) and (BRX2, (Tree, KS)) representing the MAC9s with lowest and highest power consumption, respectively.\\[-8pt]

\begin{figure}[!htb]
    \centering
    \includegraphics[width=1.0\columnwidth]{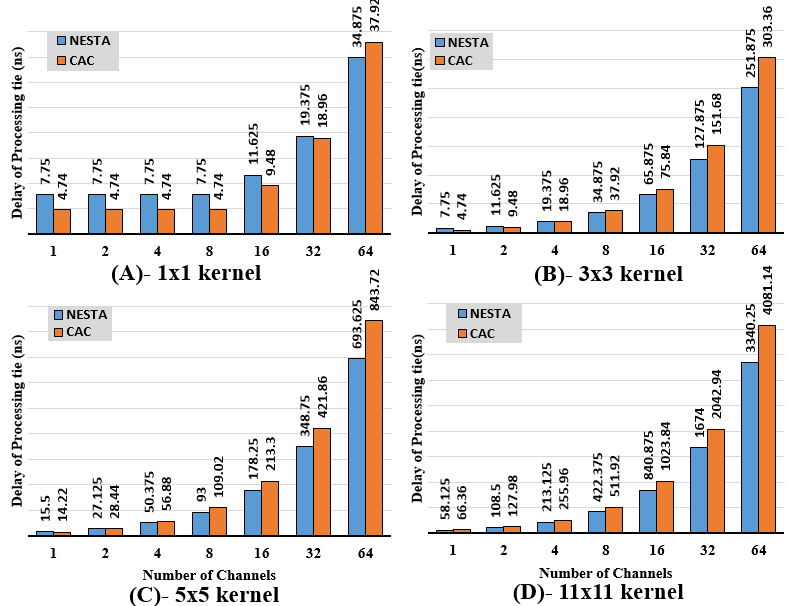}
    \caption{ Comparing the processing time of a NESTA and a MAC9 for convolutions with (A) 1x1, (B) 3x3, (C) 5x5, and (D) 11x11 kernel size when the convolution expands over multiple channels. As illustrated NESTA for larger convolutions or deeper channels becomes very more efficient.\vspace{-4mm}}
    \label{run_time_comparison}
\end{figure}

\begin{figure*}[!htb]
\vspace{-3mm}
    \centering
    \includegraphics[width=2.1\columnwidth]{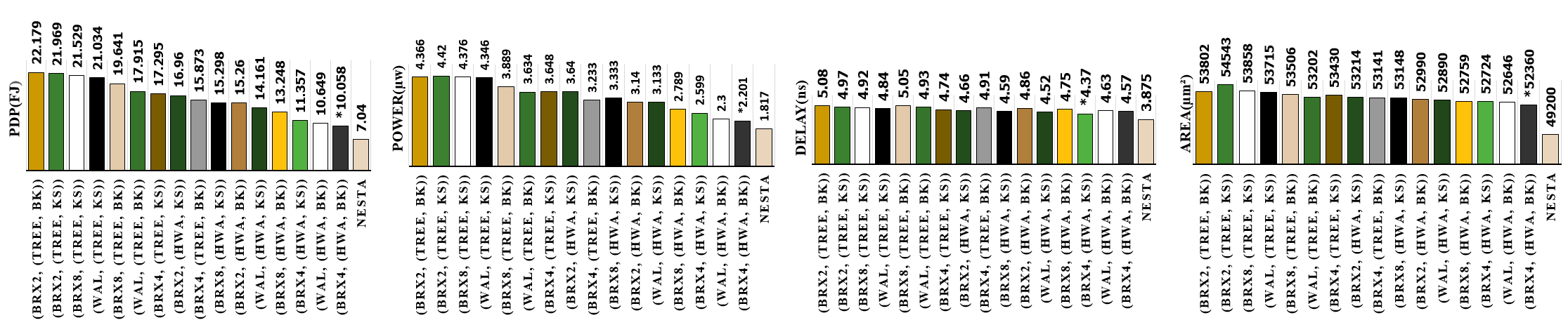} 
    \vspace{-4mm}
    \caption{Area, Delay, Power, and PDP comparison between NESTA and MAC9s constructed using fast adders and multipliers. The star identifies the best MAC9 in each category.}
    \vspace{-2mm}
    \label{hw-eval-ppa}
\end{figure*}

\textbf{Performance:} In terms of delay, the delay of NESTA is better than all other MAC9 flavors. For example, the delay of NESTA is 23.7\% and 11.3\% better than (BRX2, (Tree, BK)) and (BRX4, (HWA, KS)) as the slowest and fastest MAC9s in Fig. \ref{hw-eval-ppa}. However, the reduction in the delay comes with a catch; When NESTA process the last batch of inputs of the last channel, it has to take one extra cycle to perform the final addition. Hence, energy efficiency becomes a function of the number of processed input batches. This tradeoff is illustrated in Fig.\ref{run_time_comparison}. The larger the number of input channels, the smaller the overhead of one extra cycle for the final addition. As illustrated in Fig. \ref{run_time_comparison}, NESTA becomes more efficient if the number of processed input batches is greater than 64, 8, 2, 1 for kernel size 1x1, 3x3, 5x5, 11x11 respectively. \\[-8pt]

\textbf{Area:} Figure \ref{hw-eval-ppa} captures the PPA comparison of NESTA with various flavors of MAC9s. As illustrated, NESTA is implemented in a smaller area. The area saving is between 6\% to 9\% when NESTA is compared with (BRX4, (HWA, BK)) and (BRX2, (Tree, KS)), which are the smallest and largest MAC9s, respectively. \\

\begin{figure*}[!hbt]
    \centering
    \includegraphics[width=1.4\columnwidth]{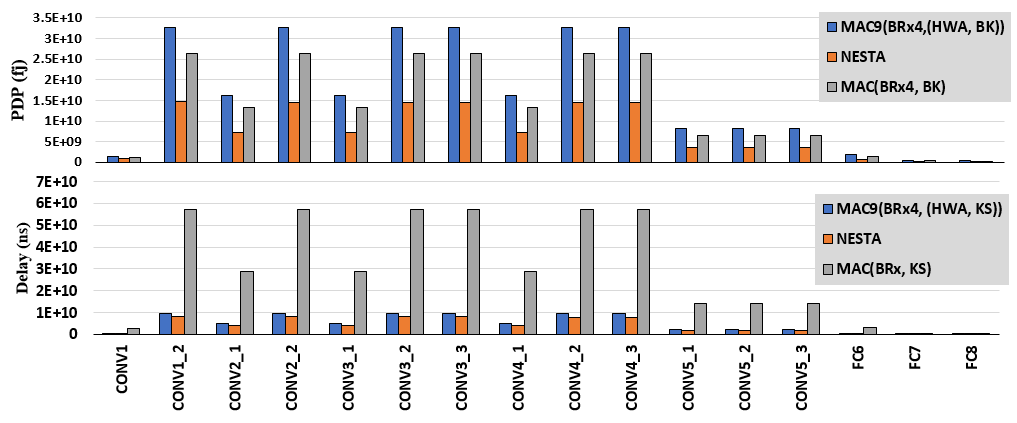}
        \vspace{-5mm}
    \caption{Breakdown of delay and energy consumption of each layer of VGG \cite{vgg} when processed by a single MAC, a single MAC9 or a single NESTA core. A linear increase in the number of cores linearly reduces the processing time.}
    \label{vgg}
\end{figure*}

\begin{figure}[!hbt]
    \centering
    \includegraphics[width=0.7\columnwidth]{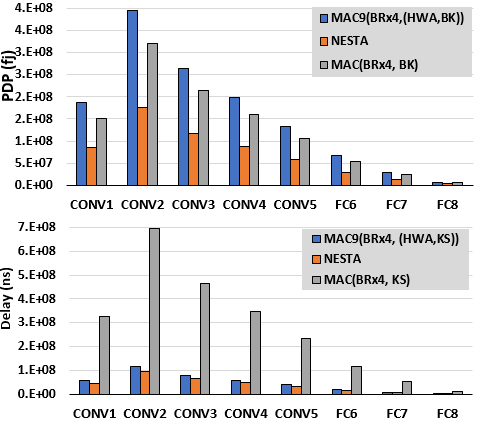}
    \vspace{-2mm}
    \caption{Breakdown of delay and energy consumption of each layer of AlexNet \cite{alexnet} when processed by a Neural engine composed of MACs, MAC9s or NESTA cores.}
    \label{alexnet}
\end{figure}

\textbf{PDP:} Considering that NESTA has lower delay and power consumption compared to other MAC9s, the PDP savings for NESTA is even more significant. According to Fig. \ref{hw-eval-ppa}, NESTA reduces the PDP by 30\% to 67\% when compared to (BRX4, (HWA, BK)) and (BRX2, (Tree, KS)) that have the lowest and highest PDP respectively.

\subsection{PPA efficiency: NESTA v.s. MACs} \label{MAC_PPA}

Table \ref{sgmVshm} captures the PPA metrics of various 2-input MACs and 9-input NESTA. Each single MAC has a smaller area, power, and delay compared to NESTA, however, in terms of functionally, one NESTA is equivalent to 9 MACs. Hence, For a fair comparison between NESTA and selected MACs, we compare their energy efficiency and throughput when fixing the area. For this comparison, we assume a NN accelerator assigns a fixed silicon area for instantiating 9-input NESTAs or 2-input MACs and report the improvement in throughput and energy with this constraint. Table \ref{impCAC_MAC} captures our comparison results. As illustrated, NESTA in terms of throughput (delay of processing normalized to the unit area) and energy efficiency (processing a large number of convolutions) is substantially more efficient than all MAC solutions studied.  By using NESTA as the PE solution in an accelerator, the throughput improves between 1\% to 37\%, correspond to (Brx4, BK) and (BRx2, KS) respectively, and energy efficiency improves 33\% to 78\%  when compared with NESTA-V1 and (BRx2, BK) which represent the best and worst MACs in terms of energy efficiency.

\begin{table}[h]
  \centering
  \vspace{-2mm}
    \caption{PPA comparison between various MAC flavors and NESTA-V1 and NESTA . }
    \scalebox{1}{
        \setlength\tabcolsep{2pt}
        \begin{tabular}{|c|c|c|c|c|}             \hline 
            \textbf{MAC Type}     & \textbf{Area}($\mu m^2$ )      & \textbf{Power}($\mu w$)     & \textbf{Delay}($ns$)   & \textbf{PDP}($fJ$) \tabularnewline            \hline 
            (BRx2, KS)   &  9394     & 0.612     & 3.57    & 2.24   \tabularnewline 
            (BRx2, BK)   &  9227     & 0.577     & 3.59	   & 2.13     \tabularnewline 
            (BRx8, KS)   &  8123     & 0.523     & 3.5     & 1.88   \tabularnewline 
            (BRx8, BK)   &  7929     & 0.509     & 3.55    & 1.86     \tabularnewline      
            (WAL, KS)    &  7024     & 0.533     & 3.46    & 1.84   \tabularnewline 
            (WAL, BK)    &  7876     & 0.566     & 3.21    & 1.81     \tabularnewline
            (BRx4, KS)   &  6899     & 0.480     & 3.10    & 1.48   \tabularnewline 
            (BRx4, BK)   &  6775     & 0.452     & 3.172   & 1.43     \tabularnewline        
            NESTA-V1   & 6825     & 0.442     & 2.914    & 1.287    \tabularnewline
            NESTA   & 49200     & 1.817     & 3.875    & 7.04    \tabularnewline
            \hline 
        \end{tabular}
    }
  \label{sgmVshm}
\end{table}

\begin{table}[h]
  \centering
    \caption{Percentage improvement in Throughput(left) \& energy consumption(right) when using NESTA to process 1K of different convolution size. \vspace{-2mm} }
    \scalebox{1}{
        \setlength\tabcolsep{2pt}
        \begin{tabular}{|c|c|c|c|c|c|c|c|c|c|}
        \cline{1-5} \cline{7-10} 
         \textbf{MAC Type}     & \textbf{3X3}  &  \textbf{5X5} &  \textbf{7X7} &  \textbf{11X11}  &  & \textbf{3X3}  &  \textbf{5X5} &  \textbf{7X7} &  \textbf{11X11} \tabularnewline
        \cline{1-5} \cline{7-10} 
         (BRx2, KS)   &  37     & 37   & 37  & 37&  & 65     & 62   & 62  & 64 \tabularnewline
        % \cline{1-5} \cline{7-10} 
         (BRx2, BK)   &  36     & 36  & 36  & 36&  & 78     & 76   & 76  & 77 \tabularnewline
        % \cline{1-5} \cline{7-10} 
         (BRx8, KS)   &  26     & 26   & 26  & 26&  &  58     & 55   & 54  & 57\tabularnewline
        % \cline{1-5} \cline{7-10} 
         (BRx8, BK)   &  25     & 25   & 25  & 25&  &  58   & 55   & 54  & 56\tabularnewline
        % \cline{1-5} \cline{7-10} 
         (WAL, KS)    &  13    & 13   & 13  & 13&  &  57    & 54  & 53 & 56 \tabularnewline
        % \cline{1-5} \cline{7-10} 
         (WAL, BK)    &  16     & 16  & 16  & 16 &  &  57     & 53  & 52  & 55\tabularnewline
        % \cline{1-5} \cline{7-10} 
         (BRx4, KS)   &  1      & 1    & 1   & 1&  &  47      & 43    & 42   & 45 \tabularnewline
        % \cline{1-5} \cline{7-10} 
         (BRx4, BK)   &  1      & 1   & 1   & 1&  & 45      & 41    & 40   & 43 \tabularnewline
        % \cline{1-5} \cline{7-10} 
         NETSA-V1     &  30     & 30   & 30  & 30&  &  39     & 34   & 33  & 37  \tabularnewline
        \cline{1-5} \cline{7-10} 
        \end{tabular}
        
        }
  \label{impCAC_MAC}
  \vspace{-4 mm}
\end{table}

\subsection{NESTA for Efficient CNN Processing}
In this section, we study the performance and energy consumption of a Neural Processing solution that uses 9-input NESTA, MAC9s, or 2-input MACs to process Alexnet\cite{alexnet} and VGG\cite{vgg}. In this paper, we only investigate the energy consumed for the processing the information and would address the saving (due to data reuse in NESTA) dataflow related power saving in the future work. Fig. \ref{vgg} and Fig. \ref{alexnet} capture the delay and energy consumed for processing each layer (including CONVs and FCs layers) of Alexnet \cite{alexnet} and VGG \cite{vgg} respectively. This is when the choice of processing engine is varied between MACs, MAC9s and NESTA cores. In each figure, NESTA is compared with the best choice of MAC or MAC9 for energy or delay according to the results of section \ref{MAC_PPA} and \ref{CAC_PPA}. As illustrated, MAC9 solutions are faster than MAC's but consume more power. However, NESTA outperforms both MAC9 and MAC solutions in terms of both power and delay (and PDP) when processing each layer of AlexNet or VGG.

\section{Acknowledgement}
This work was supported by the National Science Foundation (NSF) through Computer Systems Research (CSR) program under NSF award number 1718538.

\section{Conclusion}

In this paper, we introduced NESTA, a novel processing engine for efficient processing of Convolutional Neural Networks.  NESTA benefits from 1) its ability to generate temporal carry bits that could be passed to be included in the next round of computation without affecting the overall results, and 2) the utilization of a hierarchy of compressors to efficiently compute 9 multiplication and additions at the same time. When computing the convolution across multiple channels and/or larger convolution window sizes, NESTA generates an approximate sum ($S'$) and a temporal carry ($P$) in each cycle. In the last cycle, when processing the last convolution, NESTA takes an additional cycle and add the remaining carriers to the approximate sum to generate the correct output. Our post-layout simulation results report 30\% to 67\% reduction in power delay product (PDP) when NESTA is compared with various flavors of 9-input MAC units, and 33\% to 78\% reduction in PDP when compared with Neural processing engines constructed from various MAC flavors.

% \FloatBarrier
\nocite{icc}
\nocite{dc}
\nocite{pt}
\nocite{rh}

\bibliographystyle{IEEEtran}
\bibliography{main}

\end{document}